# On the Convergence and Robustness of Training GANs with Regularized Optimal Transport

**Maziar Sanjabi**
University of Southern California
sanjabi@usc.edu

**Jimmy Ba**
University of Toronto
jimmy@cs.toronto.edu

**Meisam Razaviyayn**
University of Southern California
razaviya@usc.edu

**Jason D. Lee**
University of Southern California
jasonlee@marshall.usc.edu

## Abstract

Generative Adversarial Networks (GANs) are one of the most practical methods for learning data distributions. A popular GAN formulation is based on the use of Wasserstein distance as a metric between probability distributions. Unfortunately, minimizing the Wasserstein distance between the data distribution and the generative model distribution is a computationally challenging problem as its objective is non-convex, non-smooth, and even hard to compute. In this work, we show that obtaining gradient information of the smoothed Wasserstein GAN formulation, which is based on regularized Optimal Transport (OT), is computationally effortless and hence one can apply first order optimization methods to minimize this objective. Consequently, we establish *theoretical convergence guarantee to stationarity* for a proposed class of GAN optimization algorithms. Unlike the original non-smooth formulation, our algorithm only requires solving the discriminator to approximate optimality. We apply our method to learning MNIST digits as well as CIFAR-10 images. Our experiments show that our method is computationally efficient and generates images comparable to the state of the art algorithms given the same architecture and computational power.

## 1 Introduction

Generative Adversarial Networks (GANs) have gained popularity for unsupervised learning due to their unique ability to learn the generation of realistic samples. In the absence of labels, GANs aims at finding the mapping from a known distribution, e.g. Gaussian, to an unknown data distribution, which is only represented by empirical samples. In order to measure the mapping quality, various metrics between the probability distributions have been proposed. While the original work on GANs proposed Jensen-Shannon distance [17], other works have proposed other metrics such as $f$-divergences [33]. Recently, a seminal work by [4] re-surged Wasserstein distance [39] as a metric for measuring the distance between the distributions. One major advantage of Wasserstein distance, compared to Jensen-Shannon, is its continuity. In addition, Wasserstein distance is differentiable with respect to the generator parameters *almost everywhere* [4]. As a result, it is more appealing from optimization perspective. From the generator perspective, this objective function is *not* smooth with respect to the generator parameters. As we will see in this paper, this non-smoothness results in difficulties for optimization algorithms. Hence, we propose to use a smooth surrogate for the Wasserstein distance.

The introduction of Wasserstein distance as a metric for GANs re-surged the interest in the field of optimal transport [39]. [4] provided a game-representation for their proposed Wasserstein GAN formulation based on the dual form of the resulting optimal transport problem. In this game-representation, the discriminator is comprised of a 1-Lipschitz function and aims at differentiating between real and fake samples. From optimization perspective, enforcing the Lipschitz constraint is challenging. Therefore, many heuristics, such as weight clipping [4] and gradient norm penalty [18], have been developed to impose such constraint on the discriminator. As we will see, our smoothed surrogate objective results in a natural penalizing term to softly impose various constraints such as Lipschitzness.

Studying the convergence of algorithms for optimizing GANs is an active area of research. The algorithms and analyses developed for optimizing GANs can be divided into two general categories based on the amount of effort spent on solving the discriminator problem. In the first category, which puts the same emphasis on the discriminator and the generator problem, a *simultaneous or successive* generator-discriminator (stochastic) gradient descent-ascent update is used for solving both problems. These approaches are inspired by the mirror/proximal gradient descent method which was developed for solving convex-concave games [30]. Although the GAN problem does not conform to convex-concave paradigm in general, researchers have found this procedure successful in *some special* practical GANs; and unsuccessful in some others [26]. The theoretical convergence guarantees for these methods are local and based on limiting assumptions which are typically not satisfied/verifiable in almost all practical GANs. More precisely, they either assume some (local) stability of the iterates or local/global convex-concave structure [29, 27, 11]. In all of these works, similar to our setting, some form of regularization is necessary for obtaining convergence. Compared to these methods, we do not limit the number of discriminator steps to one. But our convergence is global and only depends on the quality of the discriminator.

As opposed to the first line of work, some developed algorithm put more emphasis on the discriminator problem [28, 23, 19]. For example, [23] proved global convergence to optimality for the first order update rule on a specific problem of learning a one dimensional mixture of two Gaussians when the discriminator problem is solved to global optimality at each step. Another line of analysis, which also prioritize the discriminator problem, is based on the strategy of learning the discriminator much faster than the generator [19]. These analyses, which are inspired by the variants of the two time scale dynamic proposed by [7], do not directly require the convex-concavity of the objective function. However, they require some kind of local (global) stability which is difficult to achieve unless there is local (global) convex-concave structure. Compared to these results, our convergence analysis is agnostic to the method for solving the discriminator problem provided that the discriminator problem is solved to optimality within some accuracy. Based on our analysis, the amount of this accuracy then dictates the closeness to stationarity. Therefore, our result suggests that in order to obtain a better quality solution, it is not enough to just increase the number of steps for the generator, but one also needs to maintain high enough accuracy in the discriminator. It also suggests that the simple descent-ascent update rule might not converge– as it has also been observed before in the literature; see, e.g., [25] . Therefore, one should use algorithms similar to the two time scale approaches that give discriminator increasingly more advantage. Note that, unlike [23], we do not assume perfect discriminator which is not feasible in practice. It is also worth noting that the dual formulation of our regularized Wasserstein GAN is an unconstrained smooth convex problem in the functional domain. Therefore, it is theoretically feasible to solve it non-parametrically to any accuracy with polynomial number of steps in the functional space. To the best of our knowledge, our convergence analysis is the first result in the literature with mild assumptions that proves the global convergence to a stationary solution with polynomial number of generator steps and with approximate solutions to the discriminator at each step. Notice that this is possible thanks to the regularizer added to the discriminator problem in our formulation.

## 1.1 Related works and contributions

We study the problem of solving Wasserstein GANs from optimization perspective. In short, we solidify the intuitions that the use of regularized Wasserstein distance is beneficial in learning GANs [10, 6, 37, 14, 36] through a rigorous and novel algorithmic convergence analysis. There are three steps for obtaining such convergence guarantee.

- We prove that the *regularized* Wasserstein distance, when used in GAN problems, is smooth with respect to the generator parameters.

- We also prove that by *approximately* solving the *regularized* Wasserstein distance (discriminator steps), we can control the error in the computation of the (stochastic) gradients for the generator.

Such an error control could not be achieved for the original Wasserstein distance or other GAN formulations in general; see Proposition 3 in [8].

- Having *approximate* first order information and smoothness, we prove the convergence of vanilla stochastic gradient descent (SGD) method to a stationary solution. Our results suggests that converging to stationarity of the final solution not only depends on the number of steps in the generator, but also depends on the quality of solving the discriminator problem.

Note that our convergence result relies on the smoothness of *regularized* Wasserstein distance with respect to the *generator parameters*. The use of regularization as a means for smoothing has a long history in optimization literature [31]. In the optimal transport literature, the regularization has been used as a means to derive faster methods for computing the optimal transport. The most prominent example is *Sinkhorn distance* [10], which is based on regularizing the optimal transport problem with a negative entropy term. There are many efficient algorithms proposed for finding the Sinkhorn distance [21, 38, 3]. Recently Blondel et al. [6] noted that using strongly convex regularizers on the optimal transport problem would result in an unconstrained dual formulation which is computationally easier to solve. This unconstrained form is essential in using parametric methods, such as neural networks, for solving regularized optimal transport problems, as also observed in [37].

In [37] a regularized optimal transport with very small regularization weight is considered as an *objective* for learning GANs. Choosing a small regularization parameter is important as a strong regularization introduces bias. Although, our convergence guarantee applies to this objective, our smoothness analysis predicts that small weight regularization leads to an unstable algorithm. This fact has also been observed in our experiments. Thus, we use *Sinkhorn loss* [15, 36, 5] for which our convergence guarantee holds. We show that this loss does not introduce bias into finding the correct generative model regardless of the amount of regularization. Finally, using the insights from our theoretical analysis, we put together all the pieces of different methods for solving GANs with regularized optimal transport [37, 14, 36] to get an algorithm that is competitive both in terms of computational efficiency and quality with the state of the art methods.

## 2 Background

Given a cost function $c : \mathbb{R}^d \times \mathbb{R}^d \to \mathbb{R}$, the optimal transport cost between two distributions $\mathbf{p}$ and $\mathbf{q}$ can be defined as
$$d_c(\mathbf{p}, \mathbf{q}) = \min_{\pi \in \Pi(\mathbf{p}, \mathbf{q})} \int_{\mathcal{Y}} \int_{\mathcal{X}} \pi(x, y) c(x, y) dx dy, \tag{1}$$
where $\Pi(\mathbf{p}, \mathbf{q})$ is the set of all joint distributions having marginal distributions $\mathbf{p}$ and $\mathbf{q}$, i.e. $\int_{\mathcal{X}} \pi(x, y) dx = \mathbf{p}(y)$ and $\int_{\mathcal{Y}} \pi(x, y) dy = \mathbf{q}(x)$. Note that $\mathcal{X}$ and $\mathcal{Y}$ defines the space of all possible $x$'s and $y$'s respectively and in practice have finite supports. In these cases, integrals represent a finite sum throughout the paper.

The optimal transport cost (1) could be used as an objective for learning generative models. To be more specific, we assume that we have a base distribution $\mathbf{p}$ and among a set of parameterized family of functions $\{G_\theta, \theta \in \Theta\}$, we aim at learning a mapping $G_{\theta^*}$ such that the cost $d_c(G_{\theta^*}(\mathbf{q}), \mathbf{p})$ [1] is minimized. In other words the problem of generative model learning is
$$\min_{\theta \in \Theta} h_0(\theta) = d_c(G_\theta(\mathbf{q}), \mathbf{p}) = \min_{\pi \in \Pi(\mathbf{p}, \mathbf{q})} \int_{\mathcal{X}} \int_{\mathcal{Y}} \pi(x, y) \, c(G_\theta(x), y) \, dy \, dx. \tag{2}$$

### 2.1 Generative adversarial networks with $W_c$ objective

In [4] authors propose to use the dual formulation of the generative problem (2) as it is easier to parametrize the dual functions instead of the transport plan $\pi$. They call this formulation Wasserstein GAN (WGAN). Based on Kantorovich theorem [39] the dual form of (2) could be written as
$$\min_\theta \max_{\alpha, \beta} \mathbb{E}_{x \sim \mathbf{p}} \phi_\alpha(G_\theta(x)) - \mathbb{E}_{y \sim \mathbf{q}} \psi_\beta(y), \tag{3}$$
$$\text{s.t. } \phi_\alpha(G_\theta(x)) - \psi_\beta(y) \leq c(G_\theta(x), y), \forall (x, y)$$

---

[1] Throughout the paper we have the hidden technical assumption that $G_\theta$ is a one-to-one mapping on $\mathcal{X}$. This is a reasonable assumption since the space of $\mathcal{Y}$ is usually a low dimensional manifold in higher dimensional space which could be approximated by a mapping of low dimensional code words $x \in \mathcal{X}$. Therefore, the mappings are going to be from low dimensions to high dimensions.

where for practical considerations we have assumed that the dual functions/discriminators, $\phi$ and $\psi$, belong to the set of parametric functions with parameters $\alpha$ and $\beta$ respectively. Note that the inner problem has a constraint over the functions $\phi$ and $\psi$. In the case where $c$ is a distance, then $\phi = \psi$ and the constraint is enforcing the 1-Lipschitz constraint on the functions $\psi = \phi$ with respect to $c$. This 1-Lipschitzness is not easy to enforce. In practice, it is usually imposed heuristically by adding some regularizer [18].

### 2.2 Regularized optimal transport

For any strongly convex function $I(\pi)$, we can define the regularized optimal transport as

$$d_{c,\lambda}(\mathbf{p},\mathbf{q}) = \min_{\pi \in \Pi(\mathbf{p},\mathbf{q})} H(\pi,\theta) = \int\int \pi(x,y)\, c(x,y)\, dxdy + \lambda I(\pi). \tag{4}$$

We also define $\bar{d}_{c,\lambda}(\mathbf{p},\mathbf{q}) = d_{c,\lambda}(\mathbf{p},\mathbf{q}) - \lambda I(\pi^*)$, where $\pi^*$ is the optimal solution of (4). Note that, $\nabla_\theta d_{c,\lambda}(G_\theta(\mathbf{q}),\mathbf{p}) = \nabla_\theta \bar{d}_{c,\lambda}(G_\theta(\mathbf{q}),\mathbf{p})$. Among all strongly convex regularizers, the following two are the most popular ones [6]:

$$\text{KL: } I(\pi) = \int\int \pi(x,y) \log\left(\frac{\pi(x,y)}{\mathbf{q}(x)\mathbf{p}(y)}\right) dxdy \quad \& \quad \text{norm-2: } I(\pi) = \frac{1}{2}\int\int \frac{\pi(x,y)^2}{\mathbf{q}(x)\mathbf{p}(y)}\, dx\, dy.$$

When $c$ is a proper distance, one can show desirable properties for $d_{c,\lambda}$ and $\bar{d}_{c,\lambda}$; see [10] and Appendix A. It is also possible to prove the uniform convergence of function $d_{c,\lambda}(\mathbf{p},\mathbf{q})$ to $d_c(\mathbf{p},\mathbf{q})$ as $\lambda \to 0$ when $\mathbf{p}$ and $\mathbf{q}$ have finite support; see [6] and Appendix B. In the case of continuous distributions [37] proves a point-wise convergence between the two distances as $\lambda \to 0$.

### 2.3 Dual formulation for regularized optimal transport

The dual formulation for regularized optimal transport has also been covered in other recent works [37, 6]. Thus, we just present a summary of the results in Lemma 2.1 and highlight the important part in the remarks that follow; see Appendix D for a more comprehensive discussion.

**Lemma 2.1.** *Let $\phi(x)$ and $\psi(y)$ be the dual variables for the constraints in the regularized optimal transport problem* (4). *Let us also define the violation function $V(x,y) = \phi(x) - \psi(y) - c(x,y)$. Then, the dual of the regularized optimal transport is:*

$$d_{c,\lambda}(\mathbf{p},\mathbf{q}) = \max_{\psi,\phi}\ \mathbb{E}_{x\sim\mathbf{q}}[\phi(x)] - \mathbb{E}_{y\sim\mathbf{p}}[\psi(y)] - \mathbb{E}_{x,y\sim\mathbf{q}\times\mathbf{p}}\left[f_\lambda(V(x,y))\right], \tag{5}$$

*where $f_\lambda(v) = \frac{\lambda}{e}e^{\frac{v}{\lambda}}$ for KL regularization and $f_\lambda(v) = \frac{(v_+)^2}{2\lambda}$ for norm-2 regularization. Furthermore, given the optimal dual variables $\phi$ and $\psi$, the optimal primal transport plan could be computed as $\pi(x,y) = \mathbf{q}(x)\mathbf{p}(y)M(V(x,y))$, where $M(v) = \frac{1}{e}e^{\frac{v}{\lambda}}$ for KL regularization and $M(v) = \frac{v_+}{\lambda}$ for norm-2 regularization.*

**Remark 2.1.1.** *The dual of the regularized optimal transport is a large scale unconstrained concave maximization which can be solved as one. But it is also amenable to the use of parametric method, i.e., neural networks, for representing the dual functions.*

Note that $V(x,y)$ in Lemma 2.1 represents the amount of violation from the hard constraint in the original dual formulation (3). Therefore, by adding the regularizer in the primal, we are relaxing the hard constraint in the dual representation to a soft one in the objective function. By looking at the problem from this perspective, one can find similarities between our approach and the one in [18] where the authors drop the 1-Lipschitz constraint on the discriminator and try to softly enforce it by regularizing the objective using the Jacobian of the discriminator function.

**Remark 2.1.2.** *Lemma 2.1 also provides a mapping that translates the dual solutions $\phi$ and $\psi$ to a corresponding pseudo-transport plan*

$$\pi(x,y) = \mathbf{q}(x)\mathbf{p}(y)M(V(x,y)). \tag{6}$$

*Note that although $\pi$ may not be a feasible transport plan, (6) can be used to compute an approximate gradient of the generator problem, as discussed in Section 4.*

In what follows, we assume that at each iteration of the procedure for finding the optimal generator, we have access to an oracle which solves the resulting dual of regularized optimal transport to some predefined accuracy. We say a solution $(\phi,\psi)$ is an $\epsilon$-accurate solution for (5), if the value that it achieves is within $\epsilon$ of the optimal value for (5). Such an oracle could be realized through convex optimization methods and non-parametric representations; see Appendix D.2. However, due to practical computational barriers, we opt for parametric realizations of the oracle, i.e neural networks.

## 3 Smoothness of the generative objective

Given two fixed distributions $\mathbf{q}$ and $\mathbf{p}$, let us define $h_\lambda(\theta) = d_{c,\lambda}(G_\theta(\mathbf{q}), \mathbf{p})$. In this section we prove that $h_\lambda(\theta)$ is smooth with respect to $\theta$ in contrast to the original metric $h_0(\theta)$. This is particularly important when we only solve the inner optimal transport problem within some accuracy. Due to space limitations, we only state our result on the smoothness of $h_\lambda(\theta)$ when the regularizer is KL divergence; similar result for norm-2 regularizer can be found in Theorem E.1 in the Appendix. The only difference when changing the regularizer comes from the fact that these two regularizers are strongly convex with respect to different norms.

**Theorem 3.1.** *Assume there exist non-negative constants $L_1$ and $L_0$, such that:*

- *For any feasible $\theta_1, \theta_2$, $\sup_{x,y} \ \|\nabla_\theta c(G_{\theta_1}(x), y) - \nabla_\theta c(G_{\theta_2}(x), y)\| \leq L_1 \|\theta_1 - \theta_2\|$*
- *For any feasible $\theta_1$, $\sup_{x,y} \ \|\nabla_\theta c(G_{\theta_1}(x), y)\| \leq L_0$*
- *For any feasible $\theta_1, \theta_2$, $\sup_{x,y} \ |c(G_{\theta_1}(x), y) - c(G_{\theta_2}(x), y)| \leq L_0 \|\theta_1 - \theta_2\|$,*

*then the function $h_\lambda(\theta)$ is L-Lipschitz smooth, where $L = L_1 + \frac{L_0^2}{\lambda}$. Moreover, for any two parameters $\theta_1$ and $\theta_2$, $\|\pi^*(\theta_1) - \pi^*(\theta_2)\|_1 \leq \frac{L_0}{\lambda} \|\theta_1 - \theta_2\|$, where $\pi^*(\theta) = \arg\min_{\pi \in \Pi(\mathbf{p},\mathbf{q})} H(\pi, \theta)$.*

The proof of this theorem is inspired by [31] and is relegated to Appendix C. Note that unlike the non-regularized original formulation, small changes in $\theta$ results in a small change in the corresponding optimal transport plan in the regularized formulation. Consequently, after updating $\theta$, solving the inner problem would be easier as the optimal discriminator has not moved very far from the last iterate. It is also worth noticing that the assumptions of Theorem 3.1 is satisfied when the functions $c$ and $G$ are smooth and the domain of $x, y$ is compact.

## 4 Solving the generator problem to stationarity using first order methods

First order methods, including SGD and its variants such as Adam [20] or SVRG [2], are the workhorse for large scale optimization. These methods are built on top of an oracle that can generate a close approximation of the (stochastic) gradients.

Unfortunately, the original non-regularized GAN objective $h_0(\theta)$ is non-smooth. Moreover, it is impossible to obtain guaranteed good quality approximations of the its sub-gradients even if we solve the discriminator problem with high accuracy; see Proposition 3 in [8]. In contrast, we proved that the $h_\lambda(\theta)$ is smooth. Next we will prove that one can obtain decent quality estimates of its gradient by solving the corresponding regularized dual problem approximately.

**Theorem 4.1.** *Let $(\phi, \psi)$ be an $\epsilon$-accurate solution to the dual formulation of regularized optimal transport for a given $\theta$. Let $\pi$ be the transport plan corresponding to $(\phi, \psi)$, derived using (6). Let us also define $m(x, y) = \frac{\pi(x,y)}{\mathbf{q}(x)\mathbf{p}(y)}$ and $G = \mathbb{E}_{x,y \sim \mathbf{q} \times \mathbf{p}}[m(x,y)\nabla_\theta c(G_\theta(x), y)]$. Then,*

$$\|G - \nabla h_\lambda(\theta)\| \leq \delta = O\left(\sqrt{\frac{\epsilon}{\lambda}}\right) \tag{7}$$

See Appendix G for the proof. Note that it is possible to verify the quality of the discriminator/dual solutions. Due to the space limitation, we relegate the discussion on verification to Appendix D.3.

The above theorem guarantees that using the dual solver, we can generate approximate (stochastic) gradients for $h_\lambda(\theta)$. In other words, the discriminator steps in solving GANs could be viewed as a way of obtaining approximate gradient information for $h_\lambda(\theta)$. Using the above approximate (stochastic) gradients, one can provide algorithms with guaranteed convergence to approximate stationary solutions for GANs. We describe one such algorithm based on the vanilla mini-batch SGD and state its convergence guarantee.

**Remark 4.1.1.** *In Algorithm 1, if we define $G_t = \mathbb{E}[g_t | \pi_t, \theta_t]$, then Theorem 4.1 simply states that $\|G_t - \nabla h_\lambda(\theta_t)\| \leq \delta = O(\sqrt{\frac{\epsilon}{\lambda}})$.*

The following theorem establishes the convergence of Algorithm 1 to an approximate stationary solution of $h_\lambda$.

**Algorithm 1** Oracle based Non-Convex SGD for GANs

INPUT: $\mathbf{q}, \mathbf{p}, \lambda, S, \theta_0, \{\alpha_t > 0\}_{t=0}^{T-1}$
**for** $t = 0, \cdots, T-1$ **do**
    Call the oracle to find $\epsilon$-approximate maximizer $(\phi_t, \psi_t)$ for the dual formulation
    Sample I.I.D. points $x_t^1, \cdots, x_t^S \sim \mathbf{q}, y_t^1, \cdots, y_t^S \sim \mathbf{p}$ and compute
$$g_t = \frac{1}{S^2} \sum_{i,j} \frac{\pi_t(G_\theta(x_t^i), y_t^j)}{\mathbf{q}(x_t^i)\mathbf{p}(y_t^j)} \nabla_\theta c(G_\theta(x_t^i), y_t^j)$$
    where $\pi_t$ is computed using $(\phi_t, \psi_t)$ based on (6).
    Update $\theta_{t+1} \leftarrow \theta_t - \alpha_t g_t$
**end for**

---

**Theorem 4.2.** *Let $L$ be the Lipschitz constant of the gradient of $h_\lambda$. Set $\Delta = h_\lambda(\theta_0) - \inf_\theta h_\lambda(\theta)$ and let $G_t = \mathbb{E}[g_t | \pi_t, \theta_t]$. Furthermore, assume $\|G_t - \nabla h_\lambda(\theta_t)\| \leq \delta$ and $\mathbb{E}[\|g_t - G_t\|^2 | \pi_t, \theta_t] \leq \sigma^2$, $\forall t$.*

- *If $T < \frac{2\Delta L}{\sigma^2}$, setting $\alpha_t = \frac{1}{L}$, we have $\frac{1}{T} \sum_{t=1}^T \mathbb{E}[\|\nabla h_\lambda(\theta_t)\|^2] \leq \frac{2L\Delta}{T} + \delta^2 + \sigma^2$.*

- *If $T \geq \frac{2\Delta L}{\sigma^2}$, setting $\alpha_t = \sqrt{\frac{2\Delta}{L\sigma^2 T}}$, we have $\frac{1}{T} \sum_{t=1}^T \mathbb{E}\left[\|\nabla h_\lambda(\theta_t)\|_F^2\right] \leq \sigma \sqrt{8\frac{L\Delta}{T}} + \delta^2$.*

The proof of this theorem is inspired by [16] and is presented in Appendix H.

**Remark 4.2.1.** *The second regime in Theorem 4.2 results in the following asymptotic convergence rate of expected norm of the gradient as $T \to \infty$:*
$$\min_{t=1,\cdots,T} \mathbb{E}[\|\nabla_\theta h_\lambda(\theta_t)\|^2] \leq O\left(\sqrt{\frac{L}{T}}\right) + O\left(\frac{\epsilon}{\lambda}\right). \tag{8}$$
It is worth noting that our convergence analysis also guarantees the convergence of the algorithm in [37] for generative learning which is similar to Algorithm 1.

**Remark 4.2.2.** *When the error in gradient approximation at each step $t$ is $\delta_t$, Theorem 4.2 is still valid with $\delta^2 = \frac{1}{T} \sum_{t=1}^T \delta_t^2$. Thus, the algorithm only needs to keep the average error in solving the inner problem small enough.*

### 4.1 Sinkhorn loss: a more robust generative objective

In practice, when using regularized Wasserstein distance $h_\lambda(\theta)$ as an objective for generative models, one needs to use very small value of $\lambda$ as noted by [37]. This is because a large $\lambda$ would introduce bias into measuring the Wasserstein distance. In fact, choosing large $\lambda$ may lead to undesired solutions; see Corollary F.0.2 for an example. A naïve approach to deal with this bias is to reduce $\lambda$. However, a reduced $\lambda$ have three dire effects: (i) based on Theorem 3.1, any change in the generator parameters would result in large changes in the optimal discriminator parameters. (ii) According to (8), smaller $\lambda$ requires smaller error $\epsilon$ for solving discriminator to obtain the same convergence guarantee. Thus, solving the discriminator problem requires more effort. (iii) Lipschitz smoothness constant of $h_\lambda(\theta)$ increases with the decrease in $\lambda$. Thus, we have to choose smaller step-size, based on Theorem 4.2, which means slower convergence. In our experiments we observed that this situation worsens with the complexity and scale of the problem. A proposed solution in literature is to use Sinkhorn loss [15, 36] to reduce the bias in measuring the distance between the two distributions without reducing $\lambda$ and get an objective which is meaningful even for large values of $\lambda$. The Sinkhorn loss between two distributions $\mathbf{p}$ and $\mathbf{q}$ is defined as
$$L_\lambda(\mathbf{p}, \mathbf{q}) = 2\,\bar{d}_{c,\lambda}(\mathbf{p}, \mathbf{q}) - \bar{d}_{c,\lambda}(\mathbf{p}, \mathbf{p}) - \bar{d}_{c,\lambda}(\mathbf{q}, \mathbf{q}) \tag{9}$$
In [15] Genevay et al. proved that, when $c$ is a distance, as $\lambda \to \infty$, $L_\lambda$ converges to Maximum Mean Discrepancy Distance (MMD)[12]; and when $\lambda \to 0$, $L_\lambda$ converges to $2d_c(\mathbf{p}, \mathbf{q})$. Next, we present a result that shows the robustness of $L_\lambda$ with respect to the choice of $\lambda \in (0, \infty)$ in identifying the true generator parameters; see the proof in Appendix J.

**Lemma 4.3.** *Assume $c$ is symmetric, i.e., $c(x, y) = c(y, x)$. If there exists $\theta_0$ for which $\mathbf{q} = G_{\theta_0}(\mathbf{p})$, then $\theta_0$ is a stationary solution of $L_\lambda(G_\theta(\mathbf{p}), \mathbf{q})$. Moreover, $L_\lambda(G_{\theta_0}(\mathbf{p}), \mathbf{q}) = 0$ for any $\lambda > 0$.*

Notice that the above does not hold for $d_{c,\lambda}(G_\theta(\mathbf{p}), \mathbf{q})$ unless $\lambda \to 0$. Based on the above Lemma we opt to use the following Sinkhorn loss as our generative objective
$$\min_\theta \hat{h}_\lambda(\theta) = L_\lambda(G_\theta(\mathbf{q}), \mathbf{p}) = 2\,\bar{d}_{c,\lambda}(G_\theta(\mathbf{q}), \mathbf{p}) - \bar{d}_{c,\lambda}(G_\theta(\mathbf{q}), G_\theta(\mathbf{q})) - \bar{d}_{c,\lambda}(\mathbf{p}, \mathbf{p}) \tag{10}$$

Note that only the first two terms in $\hat{h}_\lambda(\theta)$ depend on $\theta$. To compute approximate gradients for these two terms, we need to call the discriminator oracle twice; and approximately solve $d_{c,\lambda}(G_\theta(\mathbf{q}), \mathbf{p})$ and $d_{c,\lambda}(G_\theta(\mathbf{q}), G_\theta(\mathbf{q}))$. With the returned discriminator solutions, we have two approximate gradients, one for each term. If each one of the gradients has error $\delta$, obtained by applying (7), the error in approximating the overall gradient is bounded by $\hat{\delta} = 3\,\delta$. Now if we further assume that sub-sampling would generate a stochastic gradient of variance $\sigma^2$ for each term, the overall variance of the noise in gradient would be bounded by $\hat{\sigma}^2 = 5\,\sigma^2$. We summarize the SGD based method for solving (10) in Algorithm 2 in the Appendix. With these assumptions, we can easily extend the convergence result of Theorem 4.2 to Algorithm 2.

**Corollary 4.3.1.** *By replacing $\sigma$ and $\delta$ in the convergence guarantees of Theorem 4.2 with $\hat{\sigma}$ and $\hat{\delta}$ respectively, we obtain a convergence guarantee for the SGD based method, described above, for solving the generative Sinkhorn loss optimization problem* (10).

## 5 Experiments

In this section we test a family of methods which we generally call Smoothed WGAN (SWGAN). They all use the two variants of regularized OT formulation, i.e. $h_\lambda(\theta)$ and $\hat{h}_\lambda(\theta)$, as their objective. We differentiate between the two objectives by explicitly mentioning Sinkhorn loss when it is used. We also investigate the choice of different cost functions, i.e. $L_1$ and Cosine distances. Unlike [36, 14] that solve regularized OT with a parametric approach, we use neural networks to solve the OT (discriminator) similar to [37]. In contrast with [36, 14] that use large batch-sizes to get unbiased gradient estimates, our gradient estimates are always unbiased due to the use of neural networks as discriminator. As a benchmark, we compare the SWGAN methods with the gradient penalty WGAN (WGAN-GP) [18] and other methods that uses the regularized OT objective [36, 15]. All algorithms were implemented in TensorFlow [1] and will be released after the acceptance.

### 5.1 Learning handwritten digits

In this section we apply SWGAN methods to learn handwritten digits on the MNIST data set. Our main goal is to see the effect of different choices of objective and cost function on the performance of SWGAN methods. For details of hyper parameters and networks structures see Appendix L.3.

The first and second row of Fig. 1 corresponds to SWGAN methods with $L_1$ and Cosine cost respectively. As the SWGAN formulations allows flexibility in the choice of the cost function, we apply these costs on different representation of the images. In Fig. 1 (a), (b), (e) and (f), the cost function is applied on the pixel domain (no latent representation), while in Fig. 1 (c), (d), (g) and (h) the cost is applied on a lower dimensional latent representation of the image [36, 14], parameterized by a Convolution Neural Network. As proposed by [36, 14], this latent representation could be adversarially trained to improve the quality of the final results; see Appendix L.1 for more details.

Comparing SWGAN methods with and without latent representation, we find that the ones with latent representation perform better. We believe this is due to the existence of many bad local minima in the high dimensional pixel space cost function which the algorithm cannot avoid. In contrast, the ones with lower dimensional latent representations could be easier cost functions to optimize globally. Based on Lemma 4.3, we conjecture that in these cases, ground truth is the only solution that is stationary regardless of the latent representation. Thus, updating the latent representation once in a while prevents the generative parameters from converging to local minima, i.e. over-fitting to a specific representation.

It is also interesting to note the difference between Fig. 1 (a) and (f), where (f) outperforms (a) which produces many faint images. We believe that the change of objective from regularized OT to Sinkhorn loss helps the method find a better stationary solution, which is closer to the underlying ground truth as predicted by Lemma 4.3. This difference is more pronounced in the experiments on CIFAR-10.

We have also included samples generated by other methods [14, 18] in the last row of Fig. 1. Compared to these methods, specially [14] which uses Sinkhorn algorithm to solve regularized OT, SWGAN methods are capable of generating higher quality images. We also noted that SWGAN methods qualitatively converge faster than other methods. We will formalize this comparison in the experiments on CIFAR-10 using the inception score [35].

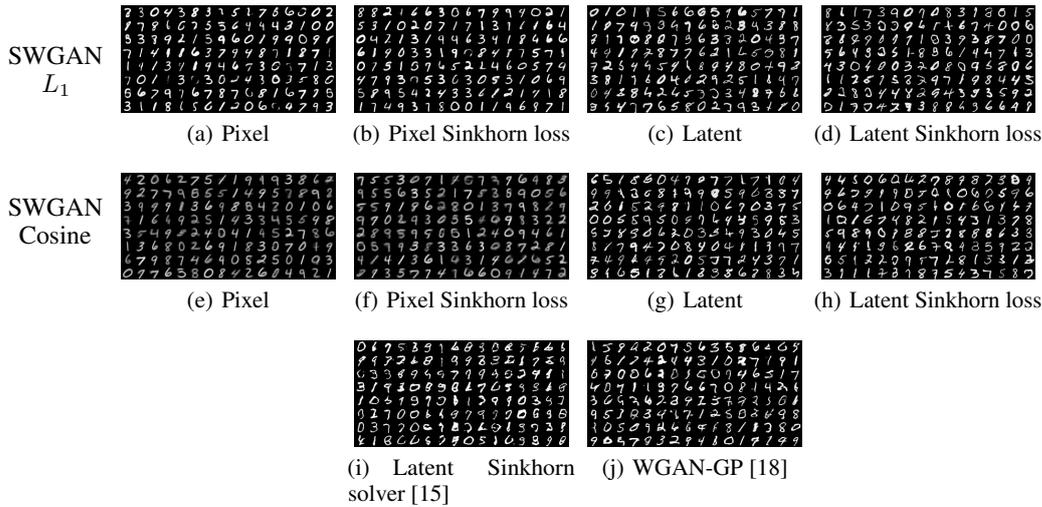

Figure 1: Generated MNIST samples using different SWGAN and benchmark methods

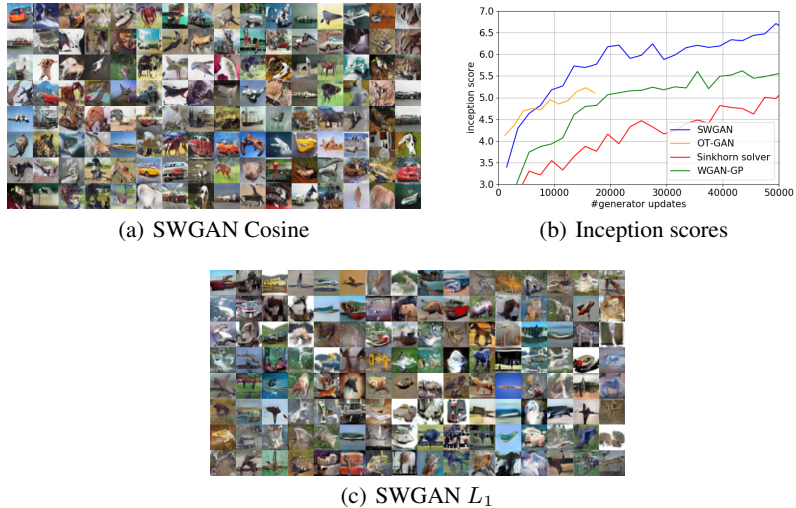

Figure 2: Generated CIFAR-10 samples and inception scores.

## 5.2 Generating tiny color images

To further investigate the performance of SWGAN, we apply it to model 32x32 color images from CIFAR-10 [22]. We compare the SWGAN approach with WGAN-GP [18], OT-GAN [36] and Sinkhorn solver [14]. All the methods are trained using the same architecture and batch-size of 150; see Appendix M for the details and a list of hyper-parameters. We use inception score [35] to compare the quality of generated samples. Learning CIFAR-10 images is a more challenging problem than MNIST; and as we predicted in Section 4.1 SWGAN methods with regularized OT objective cannot generate high quality samples, even with carefully tuned hyper-parameters; see Fig. 4 in Appendix. Due to high computational cost, we only evaluate latent Sinkhorn loss SWGAN with $L_1$ and Cosine cost on CIFAR-10. As can be seen in Fig. 2 (c), given the same architecture and computational power, the SWGAN methods have faster convergence compared to other algorithms. In Fig. 2 all the methods have been running for roughly the same amount of time. Note that OT-GAN [36] is slower as it uses two batches for each label, i.e. fake and real, and requires more computations. We also depict samples of the generated images by SWGAN methods in Fig. 2 (a) and (b). We noted that SWGAN with $L_1$ cost converges faster than the cosine distance in terms of inception scores, but the samples from the cosine distance model are more visually appealing than the ones from $L_1$.

**Algorithm 2** Oracle based Non-Convex SGD for GANs with Sinkhorn loss
___
INPUT: $\mathbf{q}, \mathbf{p}, \lambda, S, \theta_0, \{\alpha_t > 0\}_{t=0}^{T-1}$
**for** $t = 0, \cdots, T - 1$ **do**
    Call the oracle to find $\epsilon$-approximate maximizer $(\phi_t^1, \psi_t^1)$ for the dual formulation of $d_{c,\lambda}(G_\theta(\mathbf{q}), \mathbf{p})$
    Call the oracle to find $\epsilon$-approximate maximizer $(\phi_t^2, \psi_t^2)$ for the dual formulation of $d_{c,\lambda}(G_\theta(\mathbf{q}), G_\theta(\mathbf{q}))$
    Sample I.I.D. points $x_t^1, \cdots, x_t^S \sim \mathbf{q}, y_t^1, \cdots, y_t^S \sim \mathbf{p}$
    Sample I.I.D. points $\bar{x}_t^1, \cdots, \bar{x}_t^S \sim \mathbf{q}, \hat{x}_t^1, \cdots, \hat{x}_t^S \sim \mathbf{q}$
    Compute
$$g_t = \frac{2}{S^2} \sum_{i,j} \frac{\pi_t^1(G_\theta(x_t^i), y_t^j)}{\mathbf{q}(x_t^i)\mathbf{p}(y_t^j)} \nabla_\theta c(G_\theta(x_t^i), y_t^j) - \frac{1}{S^2} \sum_{i,j} \frac{\pi_t^2(G_\theta(\bar{x}_t^i), G_\theta(\hat{x}_t^i))}{\mathbf{q}(\bar{x}_t^i)\mathbf{q}(\hat{x}_t^i)} \nabla_\theta c(G_\theta(\bar{x}_t^i), G_\theta(\hat{x}_t^i)),$$
    where $\pi_t^i$ is computed using $(\phi_t^i, \psi_t^i)$ based on (6) for $i = 1, 2$.
    Update $\theta_{t+1} \leftarrow \theta_t - \alpha_t g_t$
**end for**
___

## A    Pseudo-distance property of $d_{c,\lambda}$

[10] proves that a constrained version of (4) is a pseudo-distance on the space of probabilities, i.e it is symmetric, non-negative and satisfies the triangular inequality, when $c$ is a proper distance. We prove a similar result for the regularized distance (4).

**Theorem A.1.** *If $c$ is a proper distance, then $d_{c,\lambda}$, defined in (4) is a pseudo-distance, i.e. it is non-negative, symmetric, and satisfies the triangular inequality.*

*Proof.* The proof of this theorem is very similar to [10], which proves the result for the constrained version of this objective instead of the regularized one. $d_{c,\lambda}$ is obviously symmetric and non-negative as $c$ is proper norm. In order to prove the triangular inequality, let us take three random variables $X, Y$ and $Z$. Now assume that $\pi_1$ and $\pi_2$ are the transports that achieve the minimum $d_{c,\lambda}$ between $X$ and $Z$ and $Y$ and $Z$, i.e. $\pi_1 = \arg\min_{\pi \in \Pi(X,Z)} H(\pi)$ and $\pi_2 = \arg\min_{\pi \in \Pi(Y,Z)} H(\pi)$. Now construct $\pi(x, y) = \int_z \frac{\pi_1(x,z)\pi_2(y,z)}{p(z)} dz$. It is easy to verify that $\pi(x, y) \in \Pi(X, Y)$. Moreover,

$$\int_x \int_y \pi(x,y) c(x,y)$$
$$= \int_x \int_y \int_z \frac{\pi_1(x,z)\pi_2(y,z)}{p(z)} c(x,y) \, dz \, dx \, dy$$
$$\leq \int_x \int_y \int_z \frac{\pi_1(x,z)\pi_2(y,z)}{p(z)} (c(x,z) + c(y,z)) \, dz \, dy \, dx$$
$$= \int_z \int_x \frac{\pi_1(x,z)c(x,z)}{p(z)} \underbrace{\int_y \pi_2(y,z) \, dy}_{=p(z)} dx \, dz + \int_z \int_y \frac{\pi_2(y,z)c(y,z)}{p(z)} \underbrace{\int_x \pi_1(x,z) \, dx}_{=p(z)} dy \, dz$$
$$= d_{c,\lambda}(X, Z) - \lambda I(\pi_1; X, Z) + d_{c,\lambda}(Y, Z) - \lambda I(\pi_1; Y, Z),$$

where $I(\pi; A, B)$ means the mutual information between $A$ and $B$ when their joint distribution is $\pi$. Now in order to finish the proof, we just need to prove that $I(\pi; X, Y) \leq I(\pi_1; X, Z) + I(\pi_2; Y, Z)$ in order to guarantee $d_{c,\lambda}(X, Y) \leq d_{c,\lambda}(X, Z) + d_{c,\lambda}(Y, Z)$. In fact, we prove a stronger result that $I(\pi; X, Y) \leq \min(I(\pi_1; X, Z), I(\pi_2; Y, Z))$. With some abuse of notation, let us define $p(x, y, z) = \frac{\pi(x,z)\pi(y,z)}{p(z)}$. It is obvious that $\pi(x, y) = \int_z p(x, y, z) \, dz$. Now think of $p$ as a joint distribution for $X, Y$ and $Z$. It is easy to verify that $p(x, y, z)$ could be written as

$$p(x, y, z) = p(x)p(z|x)p(y|z) = p(y)p(z|y)p(x|z). \tag{11}$$

Therefore, with joint distribution $p$, $X$ and $Y$ are independent given $Y$, i.e. $X \to Z \to Y$ and its reverse are both Markov chains. Thus, based on the data processing inequality $I(\pi; X, Y) \leq \min(I(\pi_1; X, Z), I(\pi_2; Y, Z))$. $\square$

# B Uniform convergence of $d_{c,\lambda}$ to $d_c$

We state and prove the following uniform convergence between $d_{c,\lambda}$ and $d_c$ which is very similar to the one in concurrent work [6].

**Lemma B.1.** *For any $\lambda \geq 0$,*

$$d_{c,\lambda}(G_\theta(\mathbf{p}), \mathbf{q}) - \lambda I_\theta^* \leq d_c(G_\theta(\mathbf{p}), \mathbf{q}) \leq d_{c,\lambda}(G_\theta(\mathbf{p}), \mathbf{q}), \tag{12}$$

*where $I_\theta^*$ is defined as*

$$\begin{aligned} I_\theta^* &= \min_{\pi \in \Pi(\mathbf{q},\mathbf{p})} I(\pi) \\ &\text{s.t. } \mathbb{E}_{x,y \sim \pi}[c(G_\theta(x), y)] \leq d_c(G_\theta(\mathbf{q}), \mathbf{p}). \end{aligned} \tag{13}$$

*Proof.* The proof of the rightmost inequality is a simple consequence of the fact that $I(\pi) \geq 0$. The proof of the other inequality uses the fact that $I_\theta^*$ is bounded. Thus, one can plug in the optimal transport plan into the regularized objective and get an upper-bound for the regularized optimal transport. □

In the case of discrete random variables with finite support, we can bound $I_\theta^*$ when $I$ is the KL divergence or norm-2:

**Corollary B.1.1.** *Assume $\mathbf{q}$ and $\mathbf{p}$ represent uniform discrete measures with supports of size $M$ and $N$ respectively. Let us define $K = \min(M, N)$, then*

- *if $I$ is the KL divergence, then $I_\theta^* \leq \log(K)$*
- *if $I$ is the norm-2 regularizer, then $I_\theta^* \leq \frac{K}{2}$*

*Proof.* The proof for the KL divergence is due to the fact that $I(\pi) \leq \min(\text{Entropy}(\mathbf{p}), \text{Entropy}(\mathbf{q}))$ [9]. For the case of norm-2 regularization, the proof is a simple consequence of the fact that if $\pi$ marginalizes to $\mathbf{p}$ and $\mathbf{q}$, then none of its entries could be larger than $\frac{1}{\max(M,N)}$. □

The above bound is a pessimistic bound and, in fact, the two distances might be closer in reality depending on the distributions. On the other hand, in the continuous setting, $I_\theta^*$ could be infinite. Therefore, obtaining a uniform bound on the difference between the two distances is impossible, a point-wise convergence between the two distances as $\lambda \to 0$ can be obtained [37].

# C Proof of Theorem 3.1

*Proof.* Differentiability is a consequence of the Danskin's theorem and uniqueness of the minima as $H$ is strongly convex. Moreover, for any $\theta_1$ and $\theta_2$, define $\pi_i^* = \arg\min_{\pi \in \Pi(\mathbf{p},\mathbf{q})} H(\pi, \theta_i)$, $i = 1, 2$. Due to the optimality of $\pi^*$, we have $\langle \nabla_\pi H(\theta, \pi^*(\theta)), \pi - \pi^*(\theta) \rangle \geq 0$ for all feasible $\pi$. Due to the strong convexity of $H$ with respect to $\pi$ we have:

$$H(\theta_2, \pi_2^*) \geq H(\theta_2, \pi_1^*) + \langle \nabla_\pi H(\theta_2, \pi_1^*), \pi_2^* - \pi_1^* \rangle + \frac{\lambda}{2}\|\pi_1^* - \pi_2^*\|_1^2$$

$$H(\theta_2, \pi_1^*) \geq H(\theta_2, \pi_2^*) + \underbrace{\langle \nabla_\pi H(\theta_2, \pi_2^*), \pi_1^* - \pi_2^* \rangle}_{\geq 0, \text{ due to optimality of } \pi_2^*} + \frac{\lambda}{2}\|\pi_1^* - \pi_2^*\|_1^2$$

Moreover, due to optimality of $\pi_1^*$, we have

$$\langle \nabla_\pi H(\theta_1, \pi_1^*), \pi_2^* - \pi_1^* \rangle \geq 0$$

Adding up all these inequalities, we get

$$\langle \nabla_\pi H(\theta_2, \pi_1^*) - \nabla_\pi H(\theta_1, \pi_1^*), \pi_1^* - \pi_2^* \rangle \geq \lambda \|\pi_1^* - \pi_2^*\|_1^2 \tag{14}$$

Now we use the holder inequality and get

$$\lambda \|\pi_1^* - \pi_2^*\|_1 \leq \|\nabla_\pi H(\theta_2, \pi_1^*) - \nabla_\pi H(\theta_1, \pi_1^*)\|_\infty. \tag{15}$$

Note that $\nabla_\pi(\theta, \pi) = c(G_\theta(x), y) + \lambda(1 + \log(\pi(x, y)))$. Therefore, it is easy to see that

$$\|\nabla_\pi H(\theta_2, \pi_1^*) - \nabla_\pi H(\theta_1, \pi_1^*)\|_\infty$$
$$\leq \sup_{x,y} |c(G_{\theta_1}(x), y) - c(G_{\theta_2}(x), y)|$$
$$\leq L_0 \|\theta_1 - \theta_2\|, \quad (16)$$

where the last inequality is due to the assumption. This proves the fact that $\|\pi_1^* - \pi_2^*\| \leq \frac{L_0}{\lambda}\|\theta_1 - \theta_2\|$.

To prove the Lipschitz smoothness, note that based on Danskins' theorem

$$\nabla h_\lambda(\theta) = \nabla_\theta d_{c,\lambda}(G_\theta(\mathbf{p}), \mathbf{q}) = \mathbb{E}_{x,y \sim \pi^*(\theta)}\left[\nabla_\theta c(G_\theta(x), y)\right]$$

Therefore, for any two parameters $\theta_1$ and $\theta_2$,

$$\|\nabla_\theta d_{c,\lambda}(G_{\theta_1}(\mathbf{p}), \mathbf{q}) - \nabla_\theta d_{c,\lambda}(G_{\theta_2}(\mathbf{p}), \mathbf{q})\| \leq$$
$$\|\mathbb{E}_{x,y \sim \pi_1^*}[\nabla_\theta c(G_{\theta_1}(x), y)] - \mathbb{E}_{x,y \sim \pi_2^*}[\nabla_\theta c(G_{\theta_1}(x), y)]\|$$
$$+ \|\mathbb{E}_{x,y \sim \pi_2^*}[\nabla_\theta c(G_{\theta_1}(x), y)] - \mathbb{E}_{x,y \sim \pi_2^*}[\nabla_\theta c(G_{\theta_2}(x), y)]\|$$
$$\leq L_0 \|\pi_1^* - \pi_2^*\|_1 + L_1 \|\theta_1 - \theta_2\| \leq \left(L_1 + \frac{L_0^2}{\lambda}\right)\|\theta_1 - \theta_2\|, \quad (17)$$

where the inequalities are based on triangle inequality, assumptions and the stability result we just proved respectively. □

## D  Dual formulation for regularized optimal transport

### D.1  Proof of Lemma 2.1

In this section we prove the results in Lemma 2.1 for the KL regularized optimal transport. The results for 2-norm regularized optimal transport are very similar and could be derived using the same logic.

*Proof.* Let us rewrite the primal problem for optimal transport with function $c(x, y)$ and KL regularizer.

$$\min_\pi \int_\mathcal{X} \int_\mathcal{Y} \pi(x, y) c(x, y) dx dy +$$
$$\lambda \int_\mathcal{X} \int_\mathcal{Y} \pi(x, y) \log\left(\frac{\pi(x, y)}{\mathbf{p}(y)\mathbf{q}(x)}\right) dx dy,$$
$$\text{s.t.} \int_\mathcal{X} \pi(x, y) dx = \mathbf{p}(y), \int_\mathcal{Y} \pi(x, y) dy = \mathbf{q}(x) \ \& \ \pi(x, y) \geq 0, \quad (18)$$

Now let us introduce the functions $\psi(y)$ and $-\phi(x)$ as the Lagrange multipliers for the two equality constraints respectively. Therefore, we can form the Lagrangian as follows

$$\min_{\pi(x,y) \geq 0} \int_\mathcal{X} \int_\mathcal{Y} \Big( c(x, y)\pi(x, y)$$
$$+ \lambda \pi(x, y) \log\left(\frac{\pi(x, y)}{\mathbf{p}(y)\mathbf{q}(x)}\right)$$
$$+ \psi(y)\big(\pi(x, y) - \mathbf{p}(y)\big)$$
$$- \phi(x)\big(\pi(x, y - \mathbf{q}(x))\big) \Big) dx dy \quad (19)$$

So the optimality condition for each $\pi(x, y)$ is given as

$$c(x, y) + \lambda \log\left(\pi^*(x, y)/\mathbf{q}(x)\mathbf{p}(y)\right) + \lambda + \psi(y) - \phi(x) \geq 0, \quad (20)$$

where the strict inequality holds only if $\pi^*(x, y) = 0$. But if $c(x, y) < \infty$ and $\psi$ and $\phi$ are bounded at each point, $\pi^*(x, y) > 0$. So the equality holds under the assumption that $c(x, y) < \infty$ for all $x \in \mathcal{X}, y \in \mathcal{Y}$. Based on these assumptions, the optimal $\pi^*(x, y)$ is

$$\pi^*(x, y) = \mathbf{q}(x)\mathbf{p}(y) \exp\left[\frac{\phi(x) - \psi(y) - c(x, y)}{\lambda} - 1\right] \tag{21}$$

Now if we define, $V(x, y) = \phi(x) - \psi(y) - c(x, y)$, we can rewrite the Lagrangian after plugging in the value of $\pi^*$ as

$$\max_{\psi, \phi} \ \mathbb{E}_{x \sim \mathbf{q}}[\phi(x)] - \mathbb{E}_{y \sim \mathbf{p}}[\psi(y)] - \frac{\lambda}{e} \mathbb{E}_{x, y \sim \mathbf{q} \times \mathbf{p}}\left[e^{\frac{V(x,y)}{\lambda}}\right] \tag{22}$$

$\square$

### D.2 Solving the dual: parametric vs. non-parametric

Let us focus on the case where the distributions $\mathbf{q}$ and $\mathbf{p}$ are discrete with supports of size $M$ and $N$ respectively. For simplicity, let us assume that they are empirical distributions and therefore uniform. In this case, with some abuse of notation we can define $c_{ij} = c(x_i, y_j)$, $\phi_i = \phi(x_i)$, $\psi_j = \psi(y_j)$, $\Phi = [\phi_1, \cdots, \phi_M]^T$ and $\Psi = [\psi_1, \cdots, \psi_N]^T$. Thus, the dual formulation of regularized optimal transport could be written as the following unconstrained concave maximization problem

$$F^* = \max_{\Phi, \Psi} F(\Phi, \Psi), \tag{23}$$

$$\text{where } F(\Phi, \Psi) = \frac{1}{M} \sum_i \phi_i - \frac{1}{N} \sum_i \psi_j - \frac{1}{MN} \sum_{i,j} f_\lambda(\phi_i - \psi_j - c_{ij}).$$

Here $f_\lambda$ is a convex function which depends on the regularizer and is defined in Lemma 2.1. One way to solve such a problem is to employ convex optimization algorithms. For example, one can employ a first order method, such as (stochastic) gradient descent to solve this problem [14]. Thus, one can find an $\epsilon$-accurate solution $(\hat{\Phi}, \hat{\Psi})$, such that

$$F^* - F(\hat{\Phi}, \hat{\Psi}) \leq \epsilon, \tag{24}$$

by applying gradient descent with $O(\frac{1}{\epsilon})$ iterations [32]. While this iterative procedure on this non-parametric problem is guaranteed to converge within a known number of iterations, we cannot benefit from the relationship between $x_i$'s and $y_j$'s. This becomes specially important if $\mathbf{p}$ and $\mathbf{q}$ are empirical distributions of unknown continuous distributions. In those cases, exploiting such relationships by using parametric methods, such as neural networks, can help us obtain discriminators that generalize better. In addition, exploiting such spatial relationships could also result in lower computation and memory usage. Note that for example, each gradient computation for the convex formulation requires $O(M \times N)$ computation which would be cost prohibitive. On the other hand, the parametric approaches such as neural networks could result in much lower complexities at the cost of having little theoretical guarantee. Although, we have empirically observed that parametric methods such as neural networks are capable of solving these problems efficiently.

It is also worth mentioning that Theorem 3.1 shows certain stability of the optimal regularized transport plan under small perturbations of the parameter $\theta$. Similar result can be established for optimal dual variables $(\Phi, \Psi)$; see Appendix I. This result suggests that when performing a small update on the generator parameters, solving the new dual optimal transport problem should not be very difficult if we use a warm-start obtained from the previous dual parameters.

### D.3 Verifying the quality of the dual solutions

One advantage of having an unconstrained dual is that if $F^* - F(\Phi, \Psi) \leq \epsilon$, then we have $\|\nabla F(\Phi, \Psi)\|^2 \leq 2L_F \epsilon$, where $L_F$ is the Lipschitz constant for the gradient of $F$. In other words, if $(\Phi, \Psi)$ is approximately optimal, then $\|\nabla F(\Phi, \Psi)\|$ has to be small. This condition is also applicable in the parameterized formulation, i.e., one can look at the norm of the gradient, with respect to functional values, to decide when to stop the iterative procedure for solving the discriminator problem. However, in practice obtaining this gradient is computationally expensive. In what follows, we suggest an alternative simple verifiable necessary condition to be used as the termination criterion for the discriminator iterative solver. Notice that for any $x_i$, the gradient of $F$ with respect to the corresponding

$\phi_i$ is given by $\nabla_{\phi_i} F(\Phi, \Psi) = \mathbf{q}(x_i) - \sum_j \mathbf{q}(x_i)\mathbf{p}(y_j) m_{ij}$, where $m_{ij} = m(x_i, y_j) = \frac{\pi(x_i, y_j)}{\mathbf{q}(x_i)\mathbf{p}(y_j)}$ and $\pi$ is the pseudo transport plan generated by $(\Phi, \Psi)$ using (6). Thus, in order for the gradient to be small, we need

$$\mathbb{E}_{y_j \sim \mathbf{p}} m_{ij} \approx 1. \tag{25}$$

By repeating this argument for $y_j$, we similarly obtain $\mathbb{E}_{x_i \sim \mathbf{q}} m_{ij} \approx 1$. Hence, one probabilisticnecessary condition for these two approximate inequalities to be true is that for IID samples $(x_1, y_1), \cdots, (x_S, y_S)$, we must have

$$\frac{1}{S} \sum_{k=1}^{S} m(x_k, y_k) \approx 1. \tag{26}$$

This condition is true with high probability if sample size $S$ is large enough and $m_{ij}$'s are bounded. We can use this simple measure as a dynamic termination criterion to decide when to stop the iterative discriminator optimizer.

## E  Smoothness of 2-norm regularized objective

The following Theorem states the smoothness result for the 2-norm regularized optimal transport and its proof is very similar to the the proof of Theorem 3.1 and thus omitted.

**Theorem E.1.** *If there exists non-negative constants $L_1$, $\hat{\ell}_0$ and $p_{\max}$, such that:*

- *For any feasible $\theta_1, \theta_2$,*

$$\sup_{x,y} \|\nabla_\theta c(G_{\theta_1}(x), y) - \nabla_\theta c(G_{\theta_2}(x), y)\| \leq L_1 \|\theta_1 - \theta_2\|$$

- *For any feasible $\theta$,*

$$\sqrt{\int_{\mathcal{X}} \int_{\mathcal{Y}} \|\nabla_\theta c(G_\theta(x), y)\|^2} \leq \hat{\ell}_0$$

- *For any feasible $\theta_1, \theta_2$,*

$$\sqrt{\int_{\mathcal{X}} \int_{\mathcal{Y}} |c(G_{\theta_1}(x), y) - c(G_{\theta_2}(x), y)|^2} \leq \hat{\ell}_0 \|\theta_1 - \theta_2\|,$$

- $\sup_{x,y} \mathbf{q}(x)\mathbf{p}(y) \leq p_{\max}$

*then the 2-norm regularized objective $h_\lambda(\theta) = d_{c,\lambda}(G_\theta(\mathbf{p}), \mathbf{q})$ is L-Lipschitz smooth, where $L = L_1 + \frac{\hat{\ell}_0^2 p_{\max}}{\lambda}$. Moreover, for any two parameters $\theta_1$ and $\theta_2$, $\|\pi^*(\theta_1) - \pi^*(\theta_2)\|_2 \leq \frac{\hat{\ell}_0 p_{\max}}{\lambda} \|\theta_1 - \theta_2\|$.*

We can further specialize this theorem to the case of discrete uniform distributions. In such case, $p_{\max} = \frac{1}{MN}$.

**Remark E.1.1.** *Assume uniform discrete distributions $\mathbf{q}$ and $\mathbf{p}$ with supports of size $M$ and $N$ respectively, then under the assumptions of Theorem E.1, the 2-norm regularized function $h_\lambda(\theta)$ is $L = L_1 + \frac{\ell_0^2}{\lambda}$ Lipschitz smooth, where*

$$\ell_0 = \frac{\hat{\ell}_0}{\sqrt{MN}}.$$

*Moreover, for any two parameters $\theta_1$ and $\theta_2$, $\|\pi^*(\theta_1) - \pi^*(\theta_2)\|_2 \leq \frac{\ell_0}{\lambda\sqrt{MN}} \|\theta_1 - \theta_2\|$. As a result $\|\pi^*(\theta_1) - \pi^*(\theta_2)\|_1 \leq \frac{\ell_0}{\lambda} \|\theta_1 - \theta_2\|$.*

**Remark E.1.2.** *Note that typically $\ell_0 \ll L_0$, which is defined in Theorem 3.1. This means that the norm-2 regularized surrogate is smoother than the KL regularized one, when we use the same $\lambda$.*

# F  Example: $W_2^2$-GANs with linear generator

Let us take a deeper look at the problem for the simple case where $c(z,y) = \frac{1}{2}\|z-y\|^2$, and $X$ is an $n$-dimensional zero mean random variable where $\|X\| \leq r_x$ and $Y$ is a $d$-dimensional zero mean random variable where $\|Y\| \leq r_y$ [13]. Moreover, let us assume that the generator is linear, i.e. we have matrix $\Theta \in \mathbb{R}^{d \times n}$ where $G_\Theta(x) = \Theta\, x$ and we try to find the best linear generator that maps the distribution of $X$, i.e. $\mathbf{q}$ to the distribution of $Y$, i.e. $\mathbf{p}$. In other words, we try to solve

$$\min_\Theta d_c(\Theta \mathbf{q}, \mathbf{p}), \tag{27}$$

where

$$d_c(\Theta \mathbf{q}, \mathbf{p}) = \min_{\pi \in \Pi(\mathbf{p},\mathbf{q})} \frac{1}{2} \int_{\mathcal{X},\mathcal{Y}} \pi(x,y)\|\Theta x - y\|^2 dx dy.$$

$W_2^2$ GAN formulation could be further simplified as

$$\min_\Theta \frac{1}{2}\mathrm{Tr}(\Theta \Sigma_X \Theta^T) + \frac{1}{2}\mathrm{Tr}(\Sigma_Y) + \min_{\pi \in \Pi(\mathbf{p},\mathbf{q})} \int_{\mathcal{X},\mathcal{Y}} -x^T \Theta^T y\, \pi(x,y) dx dy, \tag{28}$$

where $\Sigma_X$ and $\Sigma_Y$ are covariances for $\mathbf{q}$ and $\mathbf{p}$ respectively. As we mentioned earlier, the problem in (28) is not convex (because the last part is the min of linear functions) nor smooth (due to the fact that optimal solution of the min might not be unique). Therefore, we can approximate it with a smooth function when we add a strongly convex regularizer to the end of the inside minimization. The following two corollaries are simple applications of our smoothness result in the case of linear generator.

**Corollary F.0.1.** *If $c(z,y) = \frac{1}{2}\|z-y\|^2$, then $h_\lambda(\Theta) = d_{c,\lambda}(\Theta \mathbf{q}, \mathbf{p})$ is differentiable. Moreover, if $\forall x \in \mathcal{X}, \|x\|_2 \leq r_x$ and $\forall y \in \mathcal{Y}, \|y\|_2 \leq r_y$ and for any $\Theta_1$ and $\Theta_2$, if $\pi_i^* = \arg\min_{\pi \in \Pi} H(\Theta_i, \pi)$, $i=1,2$, then*

$$\|\pi_1^* - \pi_2^*\|_1 \leq \frac{r_x r_y}{\lambda}\|\Theta_1 - \Theta_2\|_F. \tag{29}$$

*Moreover, $h_\lambda(\theta)$ is Lipschitz smooth with respect to Frobeneous norm with constant $L = \sigma_{\max}^2(\Sigma_X) + \frac{r_x^2 r_y^2 \sqrt{nd}}{\lambda}$.*

In addition, the following corollary could be easily obtained from the above due to the special structure of the function in the linear case.

**Corollary F.0.2.** *If $\mathbf{q}$ is a uniform distribution on the sphere of radius $r_x$, then $L = \frac{r_x^2}{n} + \frac{r_x^2 r_y^2 \sqrt{nd}}{\lambda}$. Moreover, if $\lambda \geq n r_y^2 \sqrt{nd}$, then the function $d_{c,\lambda}$ would be convex and the minimizer of it would be at $\Theta = 0$.*

Basically, the above corollary says that far away from the optimal solutions, the function has a convex behavior, while the non-convexity occurs when we get close to optimal solutions. It also states that increasing $\lambda$ would cause a bias in the obtained solution towards zero and eventually will make the solution obsolete.

# G  Proof of Theorem 4.1

*Proof.* We first prove the theorem for the 2-norm regularizer. Let us define $m(x,y) = \frac{\pi(x,y)}{\mathbf{q}(x)\mathbf{p}(y)}$. With some abuse of notation, we can see that the dual problem in this case could equivalently be written as

$$\max_{\Phi,\Psi,\pi} \frac{1}{M}\sum_i \phi_i - \frac{1}{N}\sum_j \psi_j - \frac{\lambda}{2MN}\sum_{i,j} m_{ij}^2$$

$$\text{s.t. } m_{ij} \geq 0,\ m_{ij} \geq \frac{\phi_i - \psi_j - c_{ij}}{\lambda}. \tag{30}$$

This is because by optimizing over $\pi$ when fixing $\Phi$ and $\Psi$, we get the same objective as before. Now if we have a solution $(\Phi, \Psi)$ that is $\epsilon$-accurate, we can form the triple $P = (\Phi, \Psi, m)$, where $m$ is chosen to be optimal for $(\Phi, \Psi)$ (see (6) and Lemma 2.1 for the exact expression). Obviously, this triple is $\epsilon$-accurate for the alternative optimization problem (30). We can do the same for the optimal solution $(\Phi^*, \Psi^*)$ to get the triple $P^* = (\Phi^*, \Psi^*, m^*)$, which would be optimal for (30). Now both of these triples are feasible, and the feasible set is convex, therefore the line segment between them should be feasible. As a result of the optimality of $V^*$, the directional derivative of the objective in (30) at $V^*$ along this feasible direction has to be non-positive. But this means that the inner product of the gradient of the objective with the vector $\Delta P = P - P^*$ is non-positive. In addition, note that the objective is strongly concave in $m$, as a result we have the following inequality

$$\frac{\lambda}{2MN} \sum_{i,j} (m_{ij} - m_{ij}^*)^2 \leq \epsilon. \tag{31}$$

Thus, $\|m - m^*\|_2 \leq \sqrt{\frac{2MN\epsilon}{\lambda}}$. Moreover, note that

$$\nabla_\theta h_\lambda = \mathbb{E}_{x,y \sim \mathbf{q} \times \mathbf{p}} \left[ m^*(x,y) \nabla_\theta c(G_\theta(x), y) \right] \tag{32}$$

Now we use the assumptions of Remark E.1.1 on $c$ and Holder inequality to get

$$\left\| \mathbb{E}_{x,y \sim \mathbf{q} \times \mathbf{p}} \left[ \underbrace{\frac{\pi(x,y)}{\mathbf{q}(x)\mathbf{p}(y)}}_{m(x,y)} \nabla_\theta c(G_\theta(x), y) \right] - \nabla_\theta h_\lambda \right\|$$

$$= \frac{1}{MN} \| \sum_{i,j} (m_{ij} - m_{ij}^*) \nabla_\theta c(G_\theta(x), y) \|$$

$$\leq \frac{1}{MN} \ell_0 \sqrt{MN} \|m - m^*\|_2, \text{(Assumptions of Remark E.1.1)}$$

$$\leq \ell_0 \sqrt{\frac{2\epsilon}{\lambda}} \tag{33}$$

It can also be easily inferred that

$$\|\pi - \pi^*\|_1 = \frac{1}{MN} \|m - m^*\|_1 \leq \sqrt{\frac{2\epsilon}{\lambda}}.$$

In order to prove the theorem for KL regularizer, let us define $a_{ij} = \sqrt{\pi_{ij} MN}$. Now let us re-write the dual problem for KL regularized optimal transport using this new set of variables

$$\max_{\Phi, \Psi, a} \frac{1}{M} \sum_i \phi_i - \frac{1}{N} \sum_j \psi_j - \frac{\lambda}{MN} \sum_{i,j} a_{ij}^2$$

$$\text{s.t. } a_{i,j} \geq \frac{1}{\sqrt{e}} \exp\left(\frac{\phi_i - \psi_j - c_{ij}}{2\lambda}\right). \tag{34}$$

Using the same argument as before, we can form $(\Phi, \Psi, a)$ which is $\epsilon$-optimal, $(\Phi^*, \Psi^*, a^*)$ which is optimal. Simillar to the 2-norm regularized case we have

$$\|a^* - a\|_2^2 \leq \frac{\epsilon MN}{\lambda} \tag{35}$$

Now, let us define $m_{\max} = \max_{i,j}(\max(m_{ij}, m_{ij}^*))$, where $m_{ij} = a_{ij}^2$ and $m_{ij}^* = (a_{ij}^*)^2$. If we use the concavity of the square root function, we can see that $|a_{ij} - a_{ij}^*| \geq \frac{1}{2\sqrt{m_{\max}}}|m_{ij} - m_{ij}^*|$. Therefore,

$$\|m - m^*\|_2 \leq 2\sqrt{m_{\max} MN \frac{\epsilon}{\lambda}}.$$

As a result with the same argument and assumption as before we get

$$\left\| \mathbb{E}_{x,y \sim \mathbf{q} \times \mathbf{p}} \left[ \frac{\hat{\pi}(x,y)}{\mathbf{q}(x)\mathbf{p}(y)} \nabla_\theta c(G_\theta(x), y) \right] - \nabla_\theta h_\lambda \right\| \leq \ell_0 \delta,$$

where $\delta = 2\sqrt{m_{\max} \frac{\epsilon}{\lambda}}$. Similar to the previous case, it is also easy to see that $\|\pi - \pi^*\|_1 \leq \delta$.

Note that $m_{\max} \lessapprox \min(M, N)$, as a result $\delta \lessapprox 2\sqrt{\frac{\epsilon \min(M,N)}{\lambda}}$. □

# H  Proof of Theorem 4.2

*Proof.* Let us define some notations to make the proof more readable. Let us define $\nabla_t = \nabla h_\lambda(\theta_t)$ and $\mathbb{E}(g_t|\pi_t) = G_t$. Note that based on Theorem 4.1, $\|G_t - \nabla_t\| \leq \delta$.

Due to the smoothness of $d_{c,\lambda}$, which we proved in Theorem 3.1, and the update rule $\theta_{t+1} = \theta_t - \alpha_t g_t$

$$h_\lambda(\theta_{t+1}) \leq h_\lambda(\theta_t) + \langle \nabla_t, \theta_{t+1} - \theta_t \rangle + \frac{L}{2}\|\theta_{t+1} - \theta_t\|^2$$

$$= h_\lambda(\theta_t) - \alpha_t \langle \nabla_t, g_t \rangle + \frac{L\alpha_t^2}{2}\|g_t\|^2$$

$$= h_\lambda(\theta_t) - \frac{\alpha_t}{2}\left(\|\nabla_t\|^2 + \|g_t\|^2 - \|\nabla_t - g_t\|^2\right) + \frac{L\alpha_t^2}{2}\|g_t\|^2$$

Now we replace $g_t$ with $g_t - G_t + G_t$ in all the expressions and use $\|a+b\|^2 = \|a\|^2 + \|b\|^2 + 2\langle a, b \rangle$. After re-arranging the terms we get

$$h_\lambda(\theta_{t+1}) \leq h_\lambda(\theta_t) - \frac{\alpha_t}{2}\|\nabla_t\|^2 + \left(\frac{L\alpha^2 - \alpha}{2}\right)\|G_t\|^2$$
$$+ \frac{\alpha}{2}\|\nabla_t - G_t\|^2 + \frac{L\alpha^2}{2}\|G_t - g_t\|^2 - \alpha\langle \nabla_t, g_t - G_t \rangle + L\alpha^2\langle G_t, g_t - G_t \rangle \quad (36)$$

Now we are ready to sum up the inequalities across the iterations to get

$$\frac{\alpha}{2}\sum_{t=1}^T \|\nabla_t\|^2 \leq \overbrace{h_\lambda(\theta_0) - h_\lambda(\theta_{T+1})}^{\leq \Delta}$$
$$+ \sum_{t=1}^T \left[\left(\frac{L\alpha^2 - \alpha}{2}\right)\|G_t\|^2 + \frac{\alpha}{2}\|\nabla_t - G_t\|^2 + \frac{L\alpha^2}{2}\|G_t - g_t\|^2 \right.$$
$$\left. - \alpha\langle \nabla_t, g_t - G_t \rangle + L\alpha^2\langle G_t, g_t - G_t \rangle \right] \quad (37)$$

We want to take the expectation of both sides of the inequality over the randomness in the choices of $g_t$. Let us define the history up iteration $t$ as $\xi_t = [\theta_0, \cdots, \theta_t, \pi_0, \cdots, \pi_t]$. Note that $\mathbb{E}[g_t - G_t|\xi_t] = 0$, while $\nabla_t$ and $G_t$ are fixed given $\xi_t$. Therefore, by conditioning on the history while taking expectations the last two terms in the sum would be zero. In addition, as a consequence of Theorem 4.1, $\|\nabla_t - G_t\| \leq \delta$. Moreover, we have assumed $\mathbb{E}[\|G_t - g_t\|^2|\xi_t] \leq \sigma^2$. Therefore, the final inequality after taking the expectations would be

$$\frac{\alpha}{2}\sum_{t=1}^T \mathbb{E}[\|\nabla_t\|^2] \leq \Delta + T\left(\frac{\alpha\delta^2}{2} + \frac{L\alpha^2\sigma^2}{2}\right) + \frac{L\alpha^2 - \alpha}{2}\sum_{t=1}^T \mathbb{E}[\|G_t\|^2] \quad (38)$$

Now we consider two different scenarios:

- If number of iterations $T$ is large enough, i.e. $T \geq \frac{2\Delta L}{\sigma^2}$, then by choosing $\alpha_t = \alpha = \sqrt{\frac{2\Delta}{TL\sigma^2}}$, we have that $\frac{L\alpha^2 - \alpha}{2} \leq 0$. Thus, the last sum on the right hand side of (38) non-positive. Therefore, we have

$$\frac{1}{T}\sum_{t=1}^T \mathbb{E}[\|\nabla_t\|^2] \leq \sqrt{\frac{8\Delta L\sigma^2}{T}} + \delta^2.$$

- If the number of iterations is too small, i.e. $T < \frac{2\Delta L}{\sigma^2}$, then we choose $\alpha_t = \alpha = \frac{1}{L}$. In a such a case, $\frac{L\alpha^2 - \alpha}{2} = 0$. Thus, we have

$$\frac{1}{T}\sum_{t=1}^T \mathbb{E}[\|\nabla_t\|^2] \leq \frac{2L\Delta}{T} + \delta^2 + \sigma^2$$

In this case, as $T$ cannot grow to infinity, the right hand side is bounded below by $2\sigma^2 + \delta^2$.

Note that the first regime is the interesting as it results in asymptotic convergence rate of expected norm of gradient as $T \to \infty$. Therefore, it is the one that is mentioned in the body of Theorem 4.2. □

## I  Stability of dual variables

As part of Theorem 3.1 we proved the stability of the transport plan $\pi$ under small perturbations in the generator parameters $\theta$. But there are infinitely many dual variables $(\Phi, \Psi)$ corresponding to each transport plan. Therefore, it is not completely clear if the dual variables would also be stable, i.e. while we perform small changes in the generator parameter $\theta$, how far do we need to change $(\Phi, \Psi)$ to get to an optimal one for the new parameter $\theta$. The following theorem characterizes such a stability in the case of KL regularizer.

**Theorem I.1.** *Let us assume that we use the KL regularizer with weight $\lambda$ and $\mathbf{q}$ and $\mathbf{p}$ have finite supports. For any two generator parameters $\theta, \theta'$, define $\pi = \pi^*(\theta)$ and $\pi' = \pi^*(\theta')$, and $(\Phi, \Psi)$ as a set of optimal dual parameters corresponding to $\pi$. Let us further assume that $\pi, \pi' \geq \pi_{\min}$, i.e. the transport plan probabilities are bounded away from zero. Then there exists $(\Phi', \Psi')$, which corresponds to $\pi'$ and*

$$\sqrt{\|\Phi' - \Phi\|_2^2 + \|\Psi' - \Psi\|_2^2} \leq O\left(\frac{\|\theta - \theta'\|}{\pi_{\min}}\right) \tag{39}$$

**Remark I.1.1.** *The above theorem proves that when we purturb $\theta$, finding a new dual solution, starting from an old solution for the old $\theta$ should be generally easy while $\pi_{\min}$ is bunded away from zero. But it also suggests that such a problem will get more difficult when $\pi_{\min}$ goes down, i.e. when the two distributions get closer to each other or $\lambda$ is very small.*

*Proof.* Let us define $c_{ij} = c(G_\theta(x_i), y_j)$ and $c'_{ij} = c(G_\theta(x_i), y_j)$. Also we assume that the support of $\mathbf{q}$ and $\mathbf{p}$ are of sizes $M$ and $N$ respectively. From Theorem 3.1 we know that $\|\pi - \pi'\|_1 \leq \frac{L_0}{\lambda} \|\theta - \theta'\|$. Due to the mapping (6) and strong convexity of exponential function on bounded domain we have

$$\pi_{\min} \sum_{i,j} |\phi_i - \psi_j - c_{ij} - \lambda \log(e\pi'_{ij})| \leq L_0 \|\theta - \theta'\| \tag{40}$$

We also know that $|c_{ij} - c'_{ij}| \leq L_0 \|\theta - \theta'\|$. As a result

$$\sum_{i,j} |\phi_i - \psi_j - c'_{ij} - \lambda \log(e\pi'_{ij})|$$

$$\leq \sum_{i,j} |\phi_i - \psi_j - c_{ij}| + |c_{ij} - c'_{ij}|$$

$$\leq \left(\frac{L_0}{\pi_{\min}} + L_0 MN\right) \|\theta - \theta'\|$$

$$\leq L_0(1 + \pi_{\min} MN) \frac{\|\theta - \theta'\|}{\pi_{\min}} \tag{41}$$

The above result shows that $(\Phi, \Psi)$ approximately satisfy the system linear equations, i.e.

$$\phi_i - \psi_j \approx c'_{ij} + \lambda \log(e\pi'_{ij}), \quad \forall (i, j).$$

It is obvious that any accurate solution of that system of linear equations would be an optimal dual solution corresponding to $\pi'$. To find such a solution with the minimum distance to $(\Phi, \Psi)$ we just need to project $(\Phi, \Psi)$ to the subspace of all solutions of the above system of linear equations. If we call the projected point by $(\Phi', \Psi')$, the it is obviously a set of dual variables corresponding to $\pi'$. Moreover, using basic linear algebra we can see that

$$\sqrt{\|\Phi' - \Phi\|_2^2 + \|\Psi' - \Psi\|_2^2} \leq \frac{1}{\|A\|_2} \|d\|_2, \tag{42}$$

where $\|A\|_2$ is the operator norm of the linear operator corresponding to the above linear system of equation and $\|d\|$ is the norm of the violation. From (41) we know that

$$\|d\|_2 \leq L_0(1 + \pi_{\min}MN)\frac{\|\theta - \theta'\|}{\pi_{\min}}.$$

Moreover, it can be easily verified that $\|A\|_2 = \sqrt{M+N}$ for the above linear system of equation. Plugging everything back we get

$$\sqrt{\|\Phi' - \Phi\|_2^2 + \|\Psi' - \Psi\|_2^2} \leq \frac{L_0(1 + \pi_{\min}MN)}{\sqrt{M+N}}\frac{\|\theta - \theta'\|}{\pi_{\min}}$$

Note that if $\mathbf{q}$ and $\mathbf{p}$ are uniform, then it is meaningful to assume that $\pi_{\min} = \frac{c}{MN}$ for some $c \leq 1^2$. With this conversion, the bound becomes

$$\sqrt{\|\Phi' - \Phi\|_2^2 + \|\Psi' - \Psi\|_2^2} \leq L_0 \frac{(c+1)MN}{c\sqrt{M+N}}\|\theta - \theta'\|. \tag{43}$$

$\square$

**Remark I.1.2.** *Note that in equation (43) even if we control for the size of the vectors $(\Phi, \Psi)$, the stability result is getting worse linearly by when we increase $M$ and $N$. This seems to be the downside of using the dual formulation as the dual solutions will be farther apart when the number of points increases. This is in contrary with the primal stability result in which the stability is irrespctive of the number of data points.*

## J  Proof of Lemma 4.3

*Proof.* Note that in $L_\lambda(G_\theta(\mathbf{p}), \mathbf{q})$ the last term does not have any gradient with respect to $\theta$. Thus,

$$\nabla_\theta L_\lambda(G_\theta(\mathbf{p}), \mathbf{q}) = 2\nabla_\theta \bar{d}_{c,\lambda}(G_\theta(\mathbf{p}), \mathbf{q}) - \nabla_\theta \bar{d}_{c,\lambda}(G_\theta(\mathbf{p}), G_\theta(\mathbf{p})). \tag{44}$$

When $\theta = \theta_0$, we have $G_\theta(\mathbf{p}) = G_{\theta_0}(\mathbf{p}) = \mathbf{q}$. Thus, the unique optimal transport plan for $\bar{d}_{c,\lambda}(G_\theta(\mathbf{p}), G_{\theta_0}(\mathbf{p}))$ and $\bar{d}_{c,\lambda}(G_\theta(\mathbf{p}), G_\theta(\mathbf{p}))$ are the same regardless of the choice of $\lambda$. Let us denote this transport plan by $\pi^*$. Furthermore, we have:

$$\nabla_\theta \bar{d}_{c,\lambda}(G_\theta(\mathbf{p}), G_{\theta_0}(\mathbf{p})) = \int\int J_\theta(x)\nabla_1 c(G_\theta(x), G_{\theta_0}(\hat{x}))d\pi^*,$$

$$\nabla_\theta \bar{d}_{c,\lambda}(G_\theta(\mathbf{p}), G_\theta(\mathbf{p})) = \int\int J_\theta(x)\nabla_1 c(G_\theta(x), G_\theta(\hat{x}))d\pi^* +$$
$$\int\int J_\theta(\hat{x})\nabla_2 c(G_\theta(x), G_\theta(\hat{x}))d\pi^*,$$

where $\nabla_i c$ is the gradient of $c$ with respect to its $i$-th input, $i=1,2$. Moreover, $J_\theta(x)$ is the Jacobian of $G_\theta(x)$ with respect to $\theta$. Note that $x$ and $\hat{x}$ come from the same distribution $\mathbf{p}$.

Now, note that if $c$ is symmetric, as $G_\theta(\mathbf{p}) = G_{\theta_0}(\mathbf{p})$ the transport plan $\pi^*(x, \hat{x})$ is symmetric too. As $c$ is symmetric $\nabla_1 c(y, \hat{y}) = \nabla_2 c(\hat{y}, y)$. As a result if $\theta = \theta_0$

$$\int\int J_\theta(x)\nabla_1 c(G_\theta(x), G_\theta(\hat{x}))d\pi^* = \int\int J_\theta(\hat{x})\nabla_2 c(G_\theta(x), G_\theta(\hat{x}))d\pi^* \tag{45}$$

This means that $\nabla_\theta \bar{d}_{c,\lambda}(G_\theta(\mathbf{p}), G_\theta(\mathbf{p}))\Big|_{\theta=\theta_0} = 2\nabla_\theta \bar{d}_{c,\lambda}(G_\theta(\mathbf{p}), G_{\theta_0}(\mathbf{p}))\Big|_{\theta=\theta_0}$. This directly implies that $\nabla_\theta L_\lambda(G_\theta(\mathbf{p}), \mathbf{q})\Big|_{\theta=\theta_0} = 0$. $\square$

---

[2]That is because $\pi_{ij} = \frac{1}{MN}$ is equivalent to randomly assigning the two points

## K  Recovering mixture of 2D Gaussians on a grid

We first apply SWGAN on a simple synthetic data set of mixture of 25 Gaussians proposed by [24] to show the convergence of all the modes. We first use the setup proposed in [24], which uses $d = 4$ dimensional Gaussian code; for details on the architecture of the networks and hyper-parameters see Appendix L.2. Figure 3 shows the result of running our smoothed WGAN (SWGAN) algorithm for 10,000 generator iterations. The ground truth points are depicted in red while the generated points are in blue. As can be seen in the figure, our method has perfectly recovered all the modes. As a benchmark we also included the results of running WGAN-GP algorithm [18] with a set of manually tuned hyper-parameters. Note that WGAN-GP is one of the most stable algorithms in the literature. Similar to our method, WGAN-GP softly imposes 1-Lipschitzness by regularizing the objective.

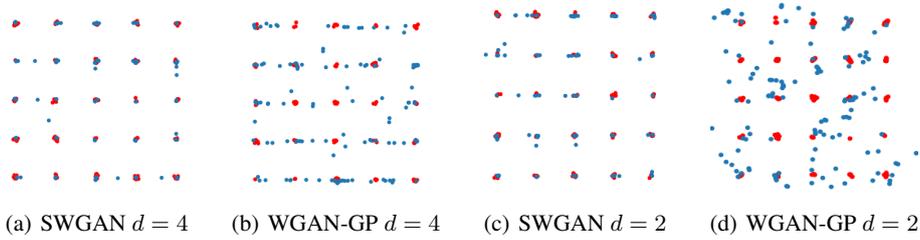

(a) SWGAN $d = 4$    (b) WGAN-GP $d = 4$    (c) SWGAN $d = 2$    (d) WGAN-GP $d = 2$

Figure 3: Learning mixture of 2-D Gaussian using random codes of dimension $d = 4$ and $d = 2$: (a) output of SWGAN with code dimension $d = 4$ after 10,000 iterations ($\approx$ 13 mins run time on a machine with one K-80 gpu) (b) output of WGAN-GP with code dimension $d = 4$ after 30,000 iterations ($\approx$ 14 mins run time on the same machine) (c) output of SWGAN with code dimension $d = 2$ after 10,000 iterations (d) output of WGAN-GP with code dimension $d = 2$ after 30,000 iterations

In order to make the problem more challenging and show the robustness of our method, we reduce the size of the random codes to $d = 2$. We use our method with the exact same hyper parameters and setup. With $d = 2$ the WGAN-GP solution quality deteriorates substantially. Figure 3 shows the results for our SWGAN as well as the best result we obtained with WGAN-GP with $d = 2$ in the same amount of time. Note that our algorithm robustly identifies all the modes even in this challenging setup for which WGAN-GP does not perform very well.

## L  Training details

### L.1  Learning data-dependent cost function

We can construct a more meaningful latent representation for images and apply $c$ in a latent space $\eta_\gamma(\cdot)$ parameterized by a set of weights $\gamma$. Note that our convergence analysis applies for a fixed representation. In practice, one can fix the representation for a few iterations and update it once in while in adversarial fashion so that it does not allow the method to converge to bad local minima. This idea has been around and many different ways for updating this representation adversarially has been proposed [5, 36, 15]. In our experiments we found that the best results would be obtained when we update the representation by applying a gradient step on parameters $\gamma$ based on the following adversariall objective:

$$\max_\gamma \bar{L}_\lambda(\gamma) = 2d_{c,\lambda}(\eta_\gamma(G_\theta(\mathbf{p})), \eta_\gamma(\mathbf{q})) + d_{c,\lambda}(\eta_\gamma(G_\theta(\mathbf{p})), \eta_\gamma(G_\theta(\mathbf{p}))) + d_{c,\lambda}(\eta_\gamma(\mathbf{q}), \eta_\gamma(\mathbf{q})). \tag{46}$$

This heuristic adversarial training of the data-dependent cost functions promotes the mapping to be more informative in differentiating between samples in the original/fake distribution as well as between fake and real distributions. We believe this would lead to learning more meaningful representations. The standard training procedure for GAN is to alternate the discriminator and the generator update. When learning the additional data-dependent $c(\eta_\gamma(x), \eta_\gamma(y))$, we update the cost function once every few generator updates. In our experiment, we used a ratio of 5 generator updates to one cost function update.

Table 1: Hyper-parameters for training mixture of 2D Gaussians.

| | |
|---|---:|
| $\lambda$ | 0.01 |
| GEN. LEARNING RATE | 0.003 |
| BATCH SIZE | $128 \times 128$ |
| DISC. LEARNING RATE | 0.001 |
| REGULARIZER | 2-NORM |
| ADAM PARAMETERS | $\beta_1 = 0.5, \beta_2 = 0.9$ |
| #DISC ITERS/GEN ITER | 20 |
| #GEN ITERS | 10,000 |
| $c(x,y)$ | $\|x - y\|_1$ |

Table 2: Hyper-parameters for training digits on MNIST.

| | |
|---|---:|
| $\lambda$ | 0.5 |
| GEN. LEARNING RATE | 2E-4 |
| BATCH SIZE | 100 |
| DISC. LEARNING RATE | 2E-4 |
| ADAM PARAMETERS | $\beta_1 = 0., \beta_2 = 0.95$ |
| #DISC ITERS/GEN ITER | 5 |
| #GEN ITER/ADVERSARIAL DISTANCE | 2 |
| #GEN ITERS | 50000 |
| $c(x,y)$ | $\|x - y\|_1$ OR $1 - \frac{x^T y}{\|x\|_2 \|y\|_2}$ |

### L.2 Training details for learning mixture of 2D Gaussians

Similar to [24] we use a generator with two fully connected hidden layers, each of which with 128 neurons with $\tanh$ activation. In [24] authors propose to use a discriminator with two fully connected hidden layers of 128 neurons and ReLU activation. For WGAN-GP we use one full discriminator with hidden layers of size 128 neurons.

In each generator iteration, we performed at most 20 iterations of discriminator, while dynamically checking the optimality condition of (26) to stop the update of discriminator.

Table 1 summarizes the chosen hyper-parameters for our method.

### L.3 Training details for MNIST digits

Table 2 summarizes the list of hyper parameters used in our training for MNIST data set.

We use a model architectures similar to DCGAN [34]. A 128 dimensional standard multivariate Gaussian is passed to a fully connected layer of 4096 hidden units. This is followed by three deconvolutional layers to generate the final 28x28 image. The discriminator has three convolutional layers using 64, 128, 256 filters with stride of 2. The feature map from the last convolutional layer is then flattened out to produce the discriminator output with a linear layer. leaky ReLU non-linearity are used in both generator and discriminators. Batch normalization and Adam are used in both generator and discriminator. The structures for our generator and discriminator networks are summerized below:

- Generator: [ FC(128, 4096)-BN-ReLU-DECONV(256, 128,stride=2)-BN-ReLU-DECONV(128, 64,stride=2)-BN-ReLU-DECONV(128, 1,stride=2)-Tanh ]

- Discriminator: [ [ CONV(1, 64,stride=2)-BN-LReLU-CONV(64, 128,stride=2)-BN-LReLU-CONV(128, 256,stride=2)-BN-LN-LReLU]-FC(4096, 5)-BN-LReLU-FC(5, 1) ]

- Adversarially learnt CNN feature denoted using []

## M Training details for CIFAR

We use the following structures for our generator and discriminator networks as [36]:

Table 3: Hyper-parameters for training CIFAR-10 images.

| | |
|---|---:|
| $\lambda$ | 0.5 |
| GEN. LEARNING RATE | 2E-4 |
| BATCH SIZE | 150 |
| DISC. LEARNING RATE | 2E-4 |
| ADAM PARAMETERS | $\beta_1 = 0., \beta_2 = 0.95$ |
| #DISC ITERS/GEN ITER | 3 |
| #GEN ITER/ADVERSARIAL DISTANCE | 2 |
| #GEN ITERS | 50000 |
| $c(x,y)$ | $\|x-y\|_1$ OR $1 - \frac{x^T y}{\|x\|_2 \|y\|_2}$ |

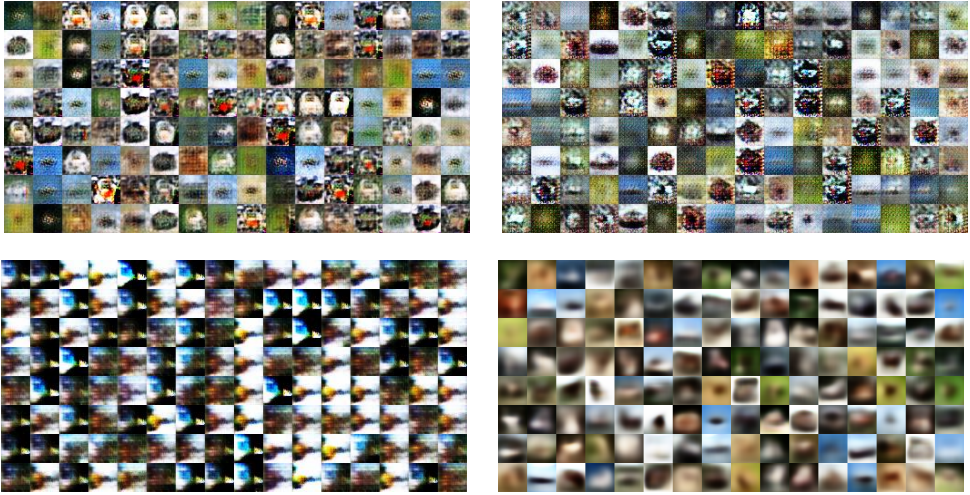

Figure 4: Results of learning CIFAR-10 images using regularized OT with manually chosen hyper-parameters and carefully annealing regularization weight: (a) These blurry samples are the best samples we could get with SWGAN with regularized OT (b) The algorithm starts becoming unstable as we anneal the regularization weight to get sharper images (c) The algorithm finally diverges and produces garbage as the discriminator loses it accuracy. (d) directly learning in the pixel space with Sinkhorn loss and fixed regularization weight.

- Generator: [ FC(128, 16384)-BN-GLU-UpSample2x-CONV(1024, 512,stride=1)-BN-GLU-UpSample2x-CONV(512, 256,stride=1)-BN-GLU-UpSample2x-CONV(256, 128,stride=1)-BN-GLU-CONV(128, 3,stride=1)-Tanh ]
- Discriminator: [ [CONV(3, 256,stride=1)-BN-CReLU-CONV(256, 512,stride=2)-BN-CReLU-CONV(64, 1024,stride=2)-BN-CReLU-CONV(128, 2048,stride=2)-BN-LN-CReLU]-FC(32768, 5)-BN-LReLU-FC(5,1) ]
- Adversarially learnt CNN feature denoted using []